\documentclass[sn-basic,Numbered]{sn-jnl}

\usepackage{graphicx}%
\usepackage{multirow}%
\usepackage{amsmath,amssymb,amsfonts}%
\usepackage{amsthm}%
\usepackage{mathrsfs}%
\usepackage[title]{appendix}%
\usepackage{xcolor}%
\usepackage{textcomp}%
\usepackage{manyfoot}%
\usepackage{booktabs}%
\usepackage{algorithm}%
\usepackage{algorithmicx}%
\usepackage{algpseudocode}%
\usepackage{listings}%
\usepackage{subcaption}
\usepackage{tikz}

\newcommand{\bftab}{\fontseries{b}\selectfont}
\begin{document}

\title[Article Title]{Language models are good pathologists: using attention-based sequence reduction and text-pre-trained transformers for efficient WSI classification}


\author*[1,2]{\fnm{Juan I.} \sur{Pisula}}\email{juan.pisula@uk-koeln.de}

\author[1,2,3]{\fnm{Katarzyna} \sur{Bozek}}\email{k.bozek@uni-koeln.de}


 \affil[1]{Institute for Biomedical Informatics, Faculty of Medicine and University Hospital Cologne, University of Cologne, Germany}

\affil[2]{Center for Molecular Medicine Cologne (CMMC), Faculty of Medicine and University Hospital Cologne, University of Cologne, Germany}

\affil[3]{Cologne Excellence Cluster on Cellular Stress Responses in Aging- Associated Diseases (CECAD), University of Cologne, Germany}


\abstract{In digital pathology, Whole Slide Image (WSI) analysis is usually formulated as a Multiple Instance Learning (MIL) problem. Although transformer-based architectures have been used for WSI classification, these methods require modifications to adapt them to specific challenges of this type of image data. Among these challenges is the amount of memory and compute required by deep transformer models to process long inputs, such as the thousands of image patches that can compose a WSI at $\times 10$ or $\times 20$ magnification. We introduce \textit{SeqShort}, a multi-head attention-based sequence shortening layer to summarize each WSI in a fixed- and short-sized sequence of instances, that allows us to reduce the computational costs of self-attention on long sequences, and to include positional information that is unavailable in other MIL approaches. Furthermore, we show that WSI classification performance can be improved when the downstream transformer architecture has been pre-trained on a large corpus of text data, and only fine-tuning less than 0.1\% of its parameters. We demonstrate the effectiveness of our method in lymph node metastases classification and cancer subtype classification tasks, without the need of designing a WSI-specific transformer nor doing in-domain pre-training, keeping a reduced compute budget and low number of trainable parameters.}

\maketitle

\section{Main}\label{sec:main} 

\begin{figure}[h]
\centering
\includegraphics[width=0.99\textwidth]{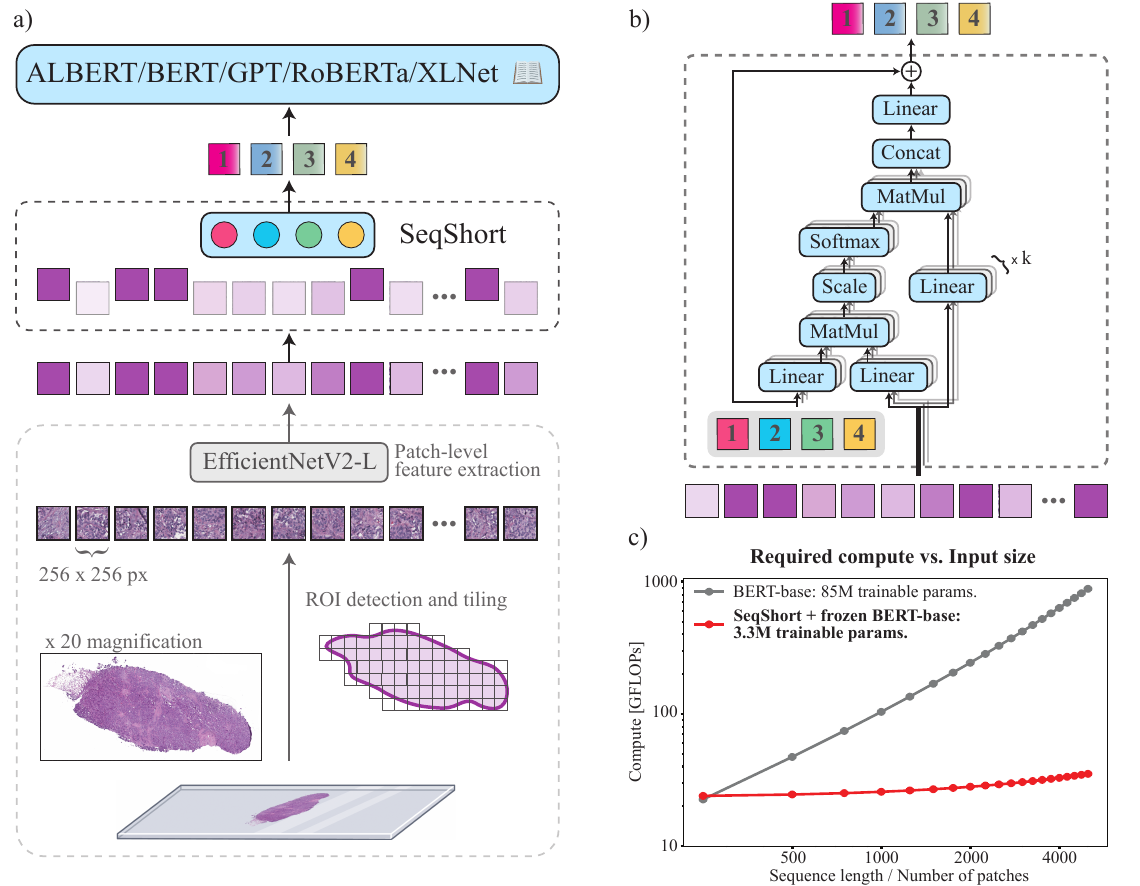}
\caption{Proposed method. a) After a typical MIL pre-processing step, our SeqShort layer summarises the amount of patches into a small, oredered sequence of feature vectors which are then classified by some popular deep transformer model that was pre-trained on an extensive text corpus. b) Detailed view of the SeqShort layer, where a set of learnable vectors query the relevant information in the WSI patches via a Multihead Attention operation. c) The computational cost of a forward pass of a deep transformer classifier is considerably reduced when our SeqShort layer is used (measured with the fvcore library by FAIR).}
\label{fig:main}
\end{figure}

Tranformers \cite{vaswani2017attention} have brought several breakthroughs to the disciplines of natural language processing (NLP) and computer vision (CV). Their capacity to link information across sequences of vector embeddings, either visual features or vectorized words, allowed to capture the structure and meaning necessary for machine translation \cite{vaswani2017attention,scaling-nmt,evolved-transf,universal-transf}, question-answering \cite{bert,albert,roberta,gpt}, image classification \cite{vit,swin,scaling-vits} and segmentation \cite{swin,setr}, and even multi-modal tasks such as text-to-image generation \cite{dalle,imagen}.

Concurrently in the field of digital pathology, the popularization of Multiple Instance Learning (MIL) \cite{mil0,mil1} approaches for Whole Slide Image (WSI) analysis allowed for the fast adoption of transformer models in this domain. By considering each WSI as a set of feature vectors of smaller tissue patches, this type of data is a natural input to transformer architectures. However, although transformer-based, these methods are always adapted to the idiosyncrasies of MIL and histopathology. Given gigapixel image size, out-of-the-box Vision Transformers (ViTs) \cite{vit} are too memory demanding. Diverse  shapes of WSIs and removal of patches consisting of background, artifacts, pen marker lines, do not allow for a straightforward implementation of local or windowed attention \cite{wsi-local0,wsi-local1} . Fixed or learnable positional embeddings are commonly removed or replaced by alternative new proposals of special positional encoding \cite{transmil,dtmil,setmil}. These challenges in WSI analysis do not allow digital pathology researchers and practitioners to take advantage of common practices and state-of-art models achieved in CV and NLP. To overcome these shortcomings, we base our work in two observations.

\textbullet\ \textbf{The redundancy of information present in full-sequence self-attention operations} can be exploited to reduce the computational cost of large inputs in deep transformer models. Wang \textit{et~al.} \cite{linformer} base their Linformer model on the observation that an attention matrix can be approximated with a matrix of lower rank. The works of Liu \textit{et~al.} \cite{swin} and Dai \textit{et~al.} \cite{funnel} propose to construct hierarchical representations instead of maintaining full-length, token-level resolution. The observations made by Clark \textit{et~al.} \cite{clark} about the importance of the [SEP] token and neighboring tokens have inspired several methods of local and sparse attention \cite{longformer, bigbird, routing, blockwise}. Comprising thousands of image patches, a WSI can be transformed to a long sequence of vector embeddings, and \textbf{we hypothesise that such findings in the transformer literature are valid to histopathology data as well}, and are necessary to allow for processing of massive in size WSIs.

\textbullet\ \textbf{Text pre-trained transformers have been proven successful in non-language related tasks.} Recent works have shown that language models pre-trained on large unstructured text corpora not only perform strongly in various downstream NLP tasks, but in several tasks outside of this domain, ranging from solving math problems \cite{math}, to zero-shot action plan generation \cite{lm-planning}. We refer to \cite{lm-planning} for an extensive enumeration of such works. In the context of CV, Ilarco \textit{et~al.} \cite{commonground} showed that text representations of frozen language models are predictive of visual representations of their corresponding object. More recently, Lu \textit{et~al.} \cite{fpt} demonstrate that pre-trained language models can perform strongly in image classification, numerical computation, and protein fold prediction when less than 0.1\% of their parameters are fine-tuned. We adhere to the claims that \textbf{language modeling pre-training can be leveraged to perform different, out-of-domain tasks,} such as WSI classification.

In this work we bridge the gap between state-of-art transformer models and digital pathology. To allow for processing of thousands of image patches from a single WSI, we propose \textit{SeqShort}, a multi-head attention (MHA) input layer that reduces long input sequences to a fixed-size short sequence that can be processed by any well known transformer model. Furthermore, we show that classification performance is increased when this technique is combined with a frozen pre-trained language model. This way, we construct a deep, yet computationally inexpensive model that requires a reduced set of trainable parameters, and performs well in digital pathology tasks, without the need of handcrafting yet another transformer variant.

\section{Results}\label{sec:results}

We showcase our results in compressing the visual information of WSIs with our sequence reduction technique in order to be classified with standard transformer models, which were pre-trained on text data in language modeling tasks. We train several architectures and find that text pre-training improves classification performance in deep transformer models. Furthermore, we show that positional information is takent into account by the transformer classifier when its input is treated as an ordered sequence, instead of an unordered collection of small image patches as commonly done in other MIL algorithms.

We then examine how our SeqShort layer works to better understand how visual information in the WSIs is aggregated. We find that only a small subset of image patches per WSI is relevant to produce our compressed sequence representations. In addition, we show that although these representations act as summaries of the WSIs, a simple extension of the attention rollout algorithm \cite{rollout} can trace the output of the transformer classifiers back to each individual image patch, providing an explanation of the classification outcome.

\subsection{WSI classification}
We measure the performance of our method on three different classification tasks: Lymph Node Metastases (LNM) classification (Normal vs Metastases); Invasive Breast Carcinoma (IBC) subtype classification (Invasive Ductal Carcinoma vs Invasive Lobular Carcinoma); and Renal Cell Carcinoma (RCC) subtype classification (Papillary Cell Carcinoma vs Chromophobe Cell Carcinoma vs Clear Cell Carcinoma). For the LNM classification task we use the dataset provided by the CAMELYON16 grand challenge (https://camelyon16.grand-challenge.org/), keeping 10\% of the training samples as a validation set, and evaluating on the grand challenge test set. For the cancer subtyping tasks, we use WSIs collected from The Cancer Genome Atlas (TCGA) (https://www.cancer.gov/tcga), and follow the same stratified 10-fold crossvalidation as \cite{clam,hipt}.

We use 256$\times$256 image patches cropped from the WSIs both at $\times10$ and $\times20$ magnification. As a data scarcity ablation, we train the models using the complete datasets or just 25\% of the samples. AUROC is used as classification performance metric. We compare our method against several weakly supervised architectures. All networks in the comparison work at a single magnification at a time, and are agnostic of how their input features vectors were produced. We use an EfficientNetV2-L \cite{efficientnet} pre-trained on ImageNet \cite{imagenet} as patch-level feature extraction network. In this experiment we use a frozen RoBERTa-base \cite{roberta} model as MIL classifier, and only fine-tune its layer normalization layer. Our LM-based model outperforms the rest of the methods in CAMELYON16 classification, and shows competitive performance in cancer subtype classification, without being designed for such digital pathology tasks, and having been pre-trained with data of a different modality (Table \ref{tab:results}).

\subsection{Language modeling pre-training improves WSI classification}
We compare the performance of different frozen text-pre-trained transformers and a baseline transformer encoder trained from scratch. Given our compute and memory constrains, the SeqShort layer was required in order to train these models. All the models under test have 12 layers of 12 attention heads and 768 hidden units, resulting in comparable transformer size. The baseline model, BERT-base, and RoBERTa-base have identical architecture and only differ in text-pre-training dataset and language modeling task. This experiment was done at $\times 20$ magnification, using the IBC task dataset. Except for ALBERT-base \cite{albert}, every model outperforms the baseline (Table \ref{tab:lm-results}).

\subsection{The role of positional information}
Because of varying WSI shapes and sizes, there are no straightforward solutions for including positional information in the aggregation of instances of MIL algorithms, particularly with transformer-based approaches. With our method, since every WSI is reduced to a fixed length sequence of feature vectors, it is possible to add to them a sequence of learned positional embeddings, which is a common practice in typical transformers for CV and NLP tasks. We demonstrate that such simple operation has a positive effect on performance with a 0.017 increase in AUROC when training at $\times10$ magnification and a 0.038 increase in AUROC when training at $\times20$ magnification (Table \ref{tab:pos-emb}).

In contrast to the ViT architecture, in which the positional embeddings learn representations indicative of the 2D location of a patch in the image, in our method, each WSI is treated as an unordered bag of patches that the SeqShort layer aggregates into a new sequence. Rather than referring to the original location of the patches in a WSI, positional information in our model makes reference to the order of the sequence generated by the SeqShort layer. Our results show that indeed there is relevant information in this order that plays a role in classification. This experiment was done on our IBC classification dataset, using RoBERTa-base as the transformer classifier.


\subsection{Insights into sequence summarization}

\begin{figure}[h]
\centering
\includegraphics[width=0.99\textwidth]{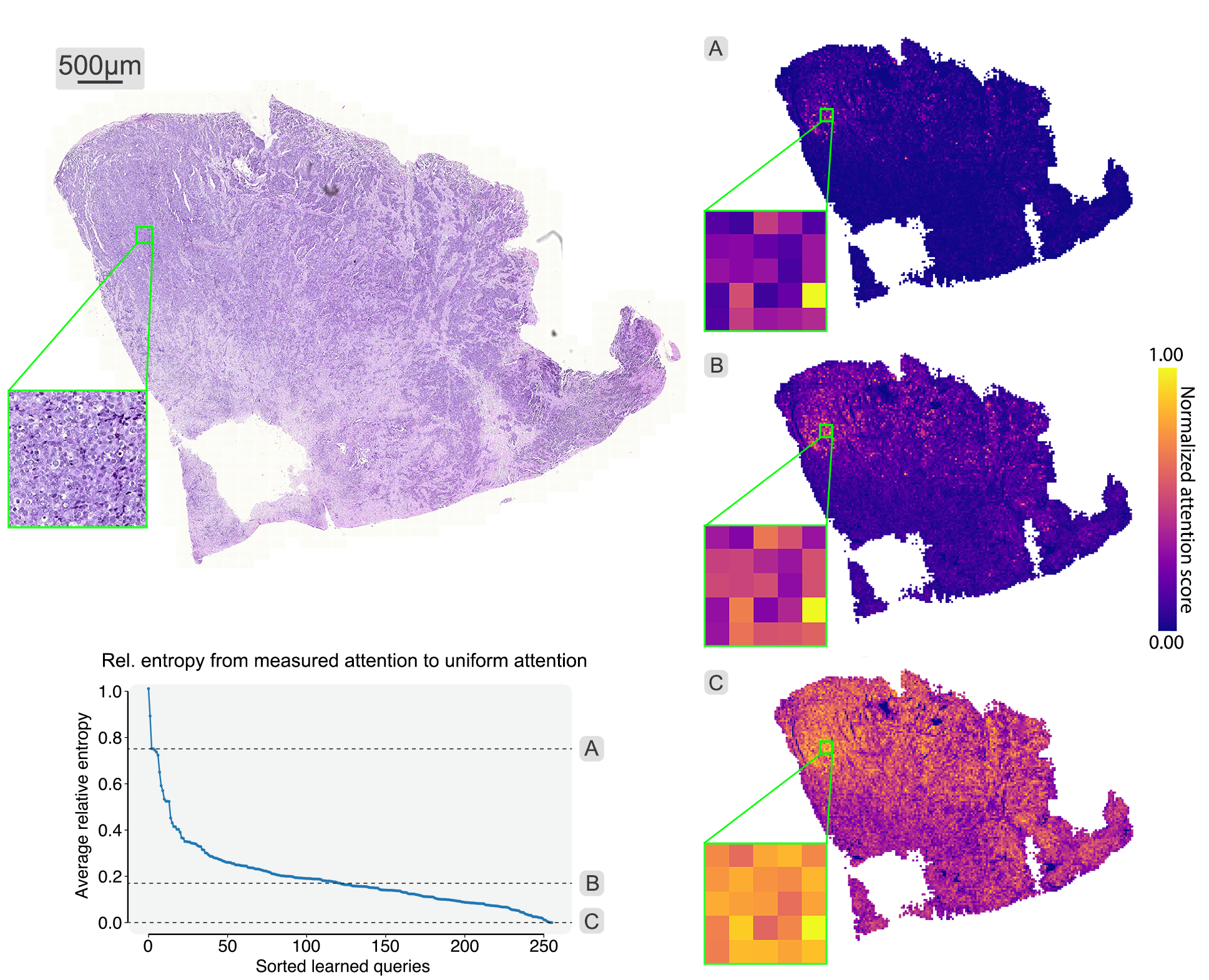}
\caption{WSI summarization inspection. A whole slide image, and attention heatmaps (A,B,C) produced by three different query vectors in SeqShort are shown. Although different queries show attentions distributed over a broader or narrower set of patches, the most important instances remain constant among the heatmaps. The bottom left plot shows the relative entropy from the attention distributions of the learned queries to uniform attention (averaged accross the dataset, and sorted for ease of visualization)}
\label{fig:entropy}
\end{figure}

We probe the SeqShort layer to examine how a WSI is summarized. We measure the relative entropy between the attention distributions produced by the different learned query vectors in SeqShort and a uniform attention distribution. Relative entropy values close to zero indicate that such queries pay overall the same amount of attention to all the input patches, whereas higher relative entropy values suggest that such queries pay more attention to a reduced subset of input instances. A plot of these values can be shown in Fig. \ref{fig:entropy}. We do this measurement with every sample of the IBC dataset test splits at $\times$20 magnification, and average the results.

Fig. \ref{fig:entropy} shows an example WSI and its respective attention heatmaps of three different query vectors with different relative entropy. Even though their attention distributions are spread over various-sized image areas, high-attention patches are shared across the heatmaps, indicating their importance over the rest.

\subsection{Explanations of classification outcome}

\begin{figure}[h]
\centering
\includegraphics[width=0.99\textwidth]{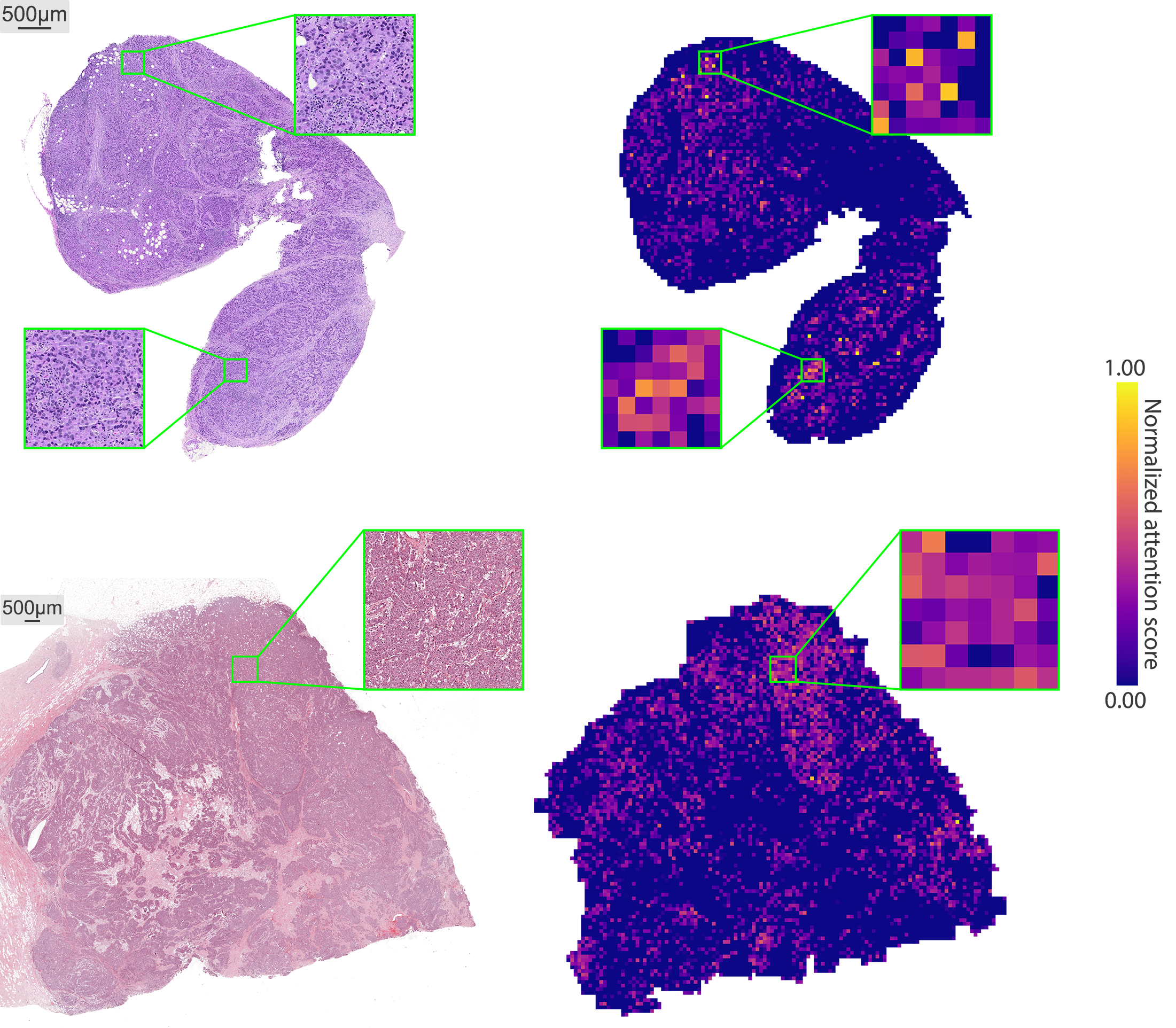}
\caption{Attention rollout heatmaps. Left: original WSIs. Right: their corresponding  attention rollout heatmaps. Although the SeqShort layer reduces the number of feature vectors that the downstream transformer has to process, it is still possible to backtrack the contribution of each individual image patch to the classification decision using the most common transformer explainability method. }
\label{fig:rollout}
\end{figure}


Attention heatmaps from the previous experiment illustrate the workings of the SeqShort layer of our model: they provide insights into how the patches of a WSI are taken into account to synthesize the intermediate output of our method.

To illustrate how the model aggregates the patch representations throughout its forward pass, we apply attention rollout \cite{rollout} to generate heatmaps that provide insights into the overall model decision process. We modify the base case of the recursive definition of attention rollout to take into consideration that SeqShort is the first layer of the complete model. Our modified attention rollout is then defined as:

\begin{equation}
  \boldsymbol{\Tilde{A}_{i}} =
    \begin{cases}
      \boldsymbol{A_{i}} \cdot \boldsymbol{\Tilde{A}_{i-1}} & \text{if $i>0$}\\
      \begin{bmatrix}
        \boldsymbol{0} \\
        \boldsymbol{A_{i}} 
      \end{bmatrix} & \text{if $i=0$},
    \end{cases}       
\end{equation}
where $\boldsymbol{A_{i}}$ is the attention matrix of layer $i$, and $\boldsymbol{0}$ is the zero vector in row space, to take into account that the [CLS] token was not present in the MHA operation of SeqShort. Example heatmaps are shown in Fig. \ref{fig:rollout}.

\section{Conclussion}\label{sec:discussion}

In this work we present a standalone layer for sequence reduction aimed to overcome common challenges in WSI classification with state-of-art transformers. Our results show that there is redundant information in the set of instances of a WSI, and that classification is possible by aggregating this set into a smaller sequence. We find that the summarized sequences have intrinsic positional information that can be leveraged by the downstream classifier. We validate our findings by examining the way our SeqShort layer aggregates the instances of a WSI.

Our sequence reduction method allows us to use powerful deep transformer architectures, previously not possible to use in digital pathology, while requiring only a fraction of their computational cost. Furthermore, we show that performance is increased when these transformers are pre-trained with language modeling task on an extensive corpus of text data. Our work supports the view that the feature representations learned by LMs can be transferred to other modalities and domains.

\section{Methods}\label{sec:methods}

\subsection{Sequence Shortening}

Existing methods \cite{funnel, hierarchical-lms, swin} for sequence reduction are not suitable for MIL WSI problems. Since there is no spatial information about instances in an unordered bag, concatenating neighboring feature vectors or taking their strided average is meaningless, as the order of the patches in a bag is arbitrary. Methods that employ a linear projection for dimensionality reduction after instance concatenation or sequence reshaping are not applicable to WSIs either, as they require a fixed and known input shape.

Here we propose using MHA for sequence shortening. Given \(\boldsymbol{X} \in \mathbb{R}^{M\times d}\) the sequence of \(M\) \(d\)-dimensional feature vectors of non-overlapping WSI tiles, we introduce our \textit{SeqShort} input layer that generates a new sequence \(\boldsymbol{X_{S}} \in \mathbb{R}^{S\times h}\) with an MHA layer:

\begin{equation}
\begin{split}
  \boldsymbol{X_{S}} & = \mbox{MHA}(Q=\boldsymbol{Q_{l}},K=\boldsymbol{X},V=\boldsymbol{X}) + \boldsymbol{Q_{l}} \\
  & = \mbox{Concat}(head_1, ..., head_k) \boldsymbol{W^O} + \boldsymbol{Q_{l}},
\end{split}
  \label{eq:seq_short_eq} 
\end{equation}
with 
\begin{equation}
    \begin{split}
        head_i = \mbox{Attention}(Q=\boldsymbol{Q_{l}}\boldsymbol{W_i^Q},K=\boldsymbol{X}\boldsymbol{W_i^K}, V=\boldsymbol{X}\boldsymbol{W_i^V}), \\
    \mbox{Attention}(Q, K, V) = \mbox{softmax}(QK^T / \sqrt{d_h})V ,
\end{split}
\end{equation}

 where \(\boldsymbol{Q_{l}} \in \mathbb{R}^{S\times h}\) is a learnable sequence of \(S\) \(h\)-dimensional query vectors, the matrices $\boldsymbol{W}$ are learnable linear projections, $d_h$ is a scaling factor commonly set as the layer's hidden dimension, and $k$ is the number of attention heads of the layer. Both \(S\) and \(h\) are hyperparameters independent of the shape of the original sequence \(\boldsymbol{X}\), and it is \(S\) which defines the output sequence length of the MHA operation in SeqShort.

This MHA operation has a sorting effect: independent of the arrangement of the patch feature vectors in $\boldsymbol{X}$, the first row of $\boldsymbol{X_{S}}$ aggregates the instances that the first query vector in $\boldsymbol{Q_{l}}$ agrees with the most; the second row of $\boldsymbol{X_{S}}$ aggregates the instances that the second query vector in $\boldsymbol{Q_{l}}$ agrees with the most, and so on. This enables to incorporate positional information in our model based on a new interpretation: instead of thinking of the original arrangement of instances in the WSI 2D space, we consider the order of the rows of $\boldsymbol{X_{S}}$ as the available positional information possible to encode.

The resulting time complexity of the MHA operation performed by our input layer is \(O(n)\) because of the fixed-size $\boldsymbol{Q_{l}}$, and since SeqShort is a single layer, the main compute load lies in the subsequent deeper transformer model in our pipeline. Although our method does not change the time complexity of the multi-head self-attention (MHSA) layers of the transformer itself, by performing sequence reduction, the amount of FLOPs it requires becomes constant with respect to the original number of WSI patches. The result is an overall considerable reduction of computational cost (Fig. \ref{fig:main}). 

\subsection{Transformer Models}

We propose the use of popular NLP transformer architectures for sequence classification. These models have not been applied before in weakly-supervised histopathology tasks given their computational cost when dealing with thousands of instances in a single WSI, whereas in NLP sequences typically comprise less than 512 tokens. The sequence shortening method that we introduce above allows us to overcome the computational cost problem.

Inspired by the success of pre-trained language models in different tasks outside NLP, we propose the use of a frozen, language-modeling pre-trained transformer as MIL classifier. This is motivated by the hypothesis that the MHSA layers of a transformer language model learn to capture the interdependencies among the elements of sequences, independent of the original data modality or domain. We follow \cite{fpt} and only fine-tune the layer normalization layers of the model, reducing the amount of trainable parameters in our transformer encoder from 85M to 36K (only 0.04\% of the total amount).

In our experiments we find that a BERT-base encoder \cite{bert} pre-trained with the masked language modeling task of \textit{Robustly optimized BERT pre-training Approach} (RoBERTa) \cite{roberta} on a corpora of more than 160GB of uncompressed text comprised by BookCorpus \cite{bookcorpus}, CC-News \cite{ccnews}, OpenWebText \cite{openweb} and Stories \cite{stories}. We discard the vocabulary embeddings lookup table of RoBERTa-base as it is not needed for weakly supervised image classification.

\subsection{Complete Pipeline}

Our pipeline is illustrated in Figure \ref{fig:main}. As a pre-processing step, we extract non-overlapping tissue tiles of $256 \times 256$ pixels from each WSI. Tissue segmentation is done as in \cite{hovernet} This step is done at $\times$20 and $\times$10 magnification for different experiments. We then generate the instance-level feature vectors using an EfficientNetV2-L \cite{efficientnet} pre-trained on ImageNet \cite{imagenet}.

The complete weakly supervised architecture that performs classification on the bag of instance vectors is composed of the SeqShort layer and a transformer language model. We set the vector embedding dimension of SeqShort to $h=768$ (the hidden dimension of the transformers under study), and $k=4$ attention heads. For the lymph node classification task we set the output length of SeqShort to $S=511$, and for the cancer subtyping tasks, this was set to $S=256$. A learnable [CLS] token is concatenated to the output of SeqShort, and added a sequence of learnable positional embeddings. The last hidden representation of [CLS] is the input of a multilayer perceptron (MLP) classification head. Altogether, our model comprises a total of 3.3M trainable parameters.

\subsection{Implementation and Training}
The method was implemented in Python, using PyTorch \cite{pytorch} as deep learning back-end. The pre-trained weights of EfficientNetV2-L and RoBERTa were downloaded from Torchvision \cite{torchvision} and HuggingFace \cite{huggingface}, respectively. Training of our models was done with the aid of PyTorch-Lightning \cite{pl}, on a single NVIDIA Tesla V100 GPU. The code of this project is available at \url{https://github.com/bozeklab/lmagp/}.

All our models were trained for 200 epochs. For the lymph node classification task the first 5 epochs were used as learning rate warm-up stage, followed by one cycle of a cosine schedule, with a maximum learning rate of $1\times 10^{-4}$, and batch size of 16. For the cancer subtyping tasks, the warm-up stage lasted 10 epochs, followed by two cycles of a cosine schedule, with a maximum learning rate of $5\times 10^{-5}$, and batch size of 32. Adam \cite{adam} was used as optimization algorithm.

\subsection{Datasets description}
\subsubsection{Lymph Node Metastases classification}
For this task we used the dataset provided by the CAMELYON16 grand challenge (https://camelyon16.grand-challenge.org/) which comprises 400 Hematoxylin and Eosin (H\&E) stained WSIs of sentinel lymph nodes of breast cancer patients, scanned by 3DHISTECH and Hamamatsu scanners at $\times40$ at the Radboud University Medical Center and the University Medical Center Utrecht, Netherlands. The grand challenge dataset is divided in a train set of 270 WSIs (160 normal slides, and 110 slides containing metastases), and a test set of 129 WSIs (80 normal slides, 49 slides containing metastases). In our experiments, we divided the provided train set in 90\%/10\% stratified splits for training and validation, respectively. 

\subsubsection{Invasive Breast Carcinoma subtype classification}
 We use a subset of 1,038 H\&E stained WSIs from the TCGA-BRCA project within The Cancer Genome Atlas repository (https://www.cancer.gov/tcga). Out of the 1,038 slides, 889 were of patients with Invasive Ductal Carcinoma, and 149 were of patients with Invasive Lobular Carcinoma. We follow the study design in \cite{clam, hipt} and evaluate the models using stratified 10-fold crossvalidation on patient level.

\subsubsection{Renal Cell Carcinoma subtype classification}
We use 918 H\&E stained WSIs of Renal Cell Carcinoma cases from the TCGA repository. Out of these samples, 289 were of Chromophobe Cell Carcinoma patients, 118 were of Papillary Cell Carcinoma patients, and 498 were of Clear Cell Carcinoma patients, coming from the TCGA-KICH, TCGA-KIRP and TCGA-KIRC projects, respectively. We follow the same study design as in the IBC subtype classification task, and evaluate the models using stratified 10-fold crossvalidation on patient level.

\section{Acknowledgments}
Both K.B. and J.I.P. were hosted by the Center for Molecular Medicine Cologne throughout this research. K.B. and J.I.P. were supported by the BMBF program Junior Group Consortia in Systems Medicine (01ZX1917B) and BMBF program for Female Junior Researchers in Artificial Intelligence (01IS20054).


\bibliography{sn-bibliography}

\newpage

\begin{table}[b]
\centering
  \caption{Performance of different MIL algorithms in the different slide-level classification tasks. Best and the second best classification results are in \textbf{bold} and \underline{underlined}, respectively.}\label{tab:results}
  
\begin{tabular}{l cc cc}
\toprule
  \multirow{2}{*}{Method} & \multicolumn{2}{c}{x10 magnification}  & \multicolumn{2}{c}{x20 magnification}  \\
                          & \multicolumn{1}{c}{25\% train set} & \multicolumn{1}{c}{100\% train set} & \multicolumn{1}{c}{25\% train set} & \multicolumn{1}{c}{100\% train set} \\ 
  \hline
  \multicolumn{5}{l}{ \multirow{2}{*}{\textbf{Lymph Node Metastases classification} }} \\ \\

  ABMIL \cite{ilse}       & 0.501 & 0.664 & \underline{0.516} & 0.616 \\
  CLAM \cite{clam}        & 0.511 & 0.692 & \underline{0.516} & 0.673 \\
  DS-MIL \cite{dsmil}     & 0.468 & \underline{0.695} & 0.441 & 0.640 \\
  TransMIL \cite{transmil}& \underline{0.529} & 0.629 & 0.470 & \underline{0.723} \\
  \bftab Ours                    & \bftab 0.627 & \bftab 0.772 & \bftab 0.642 & \bftab 0.865 \\  
  \hline

  \multicolumn{5}{l}{ \multirow{2}{*}{\textbf{Invasive Breast Carcinoma subtype classification} }} \\ \\
  ABMIL \cite{ilse}       & 0.542 ± 0.107 & 0.571 ± 0.088 & 0.551 ± 0.103 & 0.554 ± 0.107 \\
  CLAM \cite{clam}        & 0.811 ± 0.055 & 0.850 ± 0.039 & 0.697 ± 0.056 & 0.791 ± 0.082 \\
  DS-MIL \cite{dsmil}     & 0.779 ± 0.075 & 0.892 ± 0.045 & 0.711 ± 0.084 & 0.819 ± 0.082 \\
  TransMIL \cite{transmil}& \underline{0.864 ± 0.063} & \underline{0.896 ± 0.048} & \bftab 0.782 ± 0.094 & \underline{0.856 ± 0.064} \\
  \bftab Ours                    & \bftab 0.874 ± 0.052 & \bftab 0.901 ± 0.049 & \underline{0.765 ± 0.099} & \bftab 0.863 ± 0.047 \\  
  \hline
  
  \multicolumn{5}{l}{ \multirow{2}{*}{\textbf{Renal Cell Carcinoma subtype classification} }} \\ \\
  ABMIL \cite{ilse}       & 0.724 ± 0.077 & 0.795 ± 0.040 & 0.697 ± 0.077 & 0.758 ± 0.044 \\
  CLAM \cite{clam}        & \bftab 0.965 ± 0.013 & 0.969 ± 0.025 & \underline{0.961 ± 0.013} & 0.974 ± 0.010 \\
  DS-MIL \cite{dsmil}     & 0.941 ± 0.047 & 0.971 ± 0.001 & 0.926 ± 0.025 & 0.963 ± 0.001 \\
  TransMIL \cite{transmil}& \underline{0.962 ± 0.015} & \bftab 0.980 ± 0.001 & \bftab 0.971 ± 0.010 & \bftab 0.980 ± 0.001 \\
  \bftab Ours             & 0.942 ± 0.019 & \underline {0.974 ± 0.011} & 0.952 ± 0.017 & \underline{0.977 ± 0.013} \\  
\bottomrule  
\end{tabular}
\end{table}


\begin{table}[b]
\centering
\caption{Performance of different Language Models in IBC subtype classification, at $\times 20$ magnification. Baseline model is a transformer encoder of the same size but trained from scratch.}\label{tab:lm-results}
\begin{tabular}{lc}
\toprule
Language Model & {AUROC} \\
\midrule
Baseline                    & 0.784 ± 0.082 \\
XLNet-base \cite{xlnet}     & 0.819 ± 0.090 \\
GPT2-small \cite{gpt2}      & 0.827 ± 0.079 \\
BERT-base \cite{bert}       & 0.849 ± 0.058 \\
ALBERT-base \cite{albert}   & 0.747 ± 0.118 \\
RoBERTa-base \cite{roberta} & \bftab 0.863 ± 0.047 \\               
\bottomrule
\end{tabular}
\end{table}


\begin{table}[b]
\caption{Effect of including positional information on classification performance of IBC subtyping.
}\label{tab:pos-emb}
  
\begin{tabular}{lll}
\toprule
Magnification                & Pos. emebddings & \multicolumn{1}{c}{AUROC} \\ 
\midrule
\multirow{2}{*}{$\times 20$} & No              & 0.825 ± 0.052             \\
                             & \textbf{Yes}    & \bftab 0.863 ± 0.047      \\
\multirow{2}{*}{$\times 10$} & No              & 0.884 ± 0.062             \\
                             & \textbf{Yes}    & \bftab 0.901 ± 0.049      \\         
\bottomrule
\end{tabular}
\end{table}

\end{document}